\pdfoutput=1

\documentclass[11pt]{article}

\usepackage{ACL2023}
\usepackage{times}
\usepackage{latexsym}
\usepackage[T1]{fontenc}
\usepackage[utf8]{inputenc}
\usepackage{microtype}
\usepackage{inconsolata}
\usepackage{graphicx}

\title{Identification of Knowledge Neurons in Protein Language Models}

\author{Divya Nori \and Shivali Singireddy \and Marina Ten Have \\
        Massachusetts Institute of Technology, Cambridge, Massachusetts \\ divnor80@mit.edu \\ shivali5@mit.edu \\ mtenhave@mit.edu}

\begin{document}
\maketitle
\begin{abstract}

Neural language models have become powerful tools for learning complex representations of entities in natural language processing tasks. However, their interpretability remains a significant challenge, particularly in domains like computational biology where trust in model predictions is crucial. In this work, we aim to enhance the interpretability of protein language models, specifically the state-of-the-art ESM model, by identifying and characterizing "knowledge neurons" – components that express understanding of key information. After fine-tuning the ESM model for the task of enzyme sequence classification, we compare two knowledge neuron selection methods that preserve a subset of neurons from the original model. The two methods, activation-based and integrated gradient-based selection, consistently outperform a random baseline. In particular, these methods show that there is a high density of knowledge neurons in the key vector prediction networks of self-attention modules. Given that key vectors specialize in understanding different features of input sequences, these knowledge neurons could capture knowledge of different enzyme sequence motifs. In the future, the types of knowledge captured by each neuron could be characterized.

\end{abstract}

\section{Introduction}

Neural language models, such as transformers and recurrent neural networks (RNNs), have proven incredibly powerful in learning rich representations of entities for downstream natural language processing tasks. However, they are often considered ``black boxes," making it challenging to understand and interpret the representations they learn \cite{sun2021interpreting}. This lack of interpretability is an especially critical issue in the domain of computational biology, where neural language models are increasingly being used for high-liability applications such as drug discovery and disease prediction. In molecular biology, language models have broadly been applied to learn representations of proteins for downstream tasks \cite{wu2023integration}. For example, the Evolutionary Scale Model (ESM) transformer is the state-of-the-art protein language model, aiding in tasks like protein property, structure, and function prediction \cite{verkuil2022language}. However, it remains unclear how exactly ESM is able to make its predictions, decreasing the amount of trust scientists place in the model’s representations and decisions. 

The lack of interpretability in language models arises from the complexity of their architecture \cite{zafar2021lack}. These models contain numerous layers and millions of parameters, making it difficult to discern how they arrive at their predictions. Additionally, it is difficult to assess which components of a pre-trained model are most important for a downstream task. Our research project aims to address this problem of interpretability in protein language models, focusing on the ESM-2 transformer model fine-tuned for the task of enzyme sequence classification.

We evaluate two methods to reveal which components of the pre-trained model learn the most relevant information, allowing for the identification of \textit{knowledge neurons} that express understanding of key facts, such as motifs in protein sequences that are strong indicators of certain enzyme classes. In this study, we specifically evaluate activation-based and integrated gradients-based methods to label each neuron as knowledge-expressing or not. This work serves as a first step towards characterizing the components of large biological models, eventually building towards discovering which facts each neuron learns to express. By shedding light on the learned representations of protein language models, we can potentially enhance their accuracy and utility in these critical applications.

\section{Related Work}

\subsection{Language Model Interpretability}

In the past few years, a growing body of work has been devoted to analyzing the inner workings of neural network models in natural language processing. \citet{dai2021knowledge} proposed that pre-trained language models contain \textit{knowledge neurons}, proposing that certain neurons learn specific concepts, and these neurons are most important in tasks requiring expression of certain knowledge. The authors of this work investigate two methods for identifying knowledge neurons: analyzing neuron activations and integrated gradients. After identifying knowledge neurons in the Bidirectional Encoder Representations from Transformers (BERT) model using both methods, the authors of this work show that a network consisting of just the knowledge neurons performs better than a random subset of the original model, where all submodels are the same size \cite{devlin2018bert}.

A few papers have attempted to deconstruct model architectures through analysis of neuron activations, based on the hypothesis that neurons with higher activation values correlate with expression of knowledge \cite{rethmeier2020tx}, \cite{meng2022locating}. An activation value refers to the output produced by a neuron. Selecting knowledge neurons by activation magnitude is a reasonable method to evaluate because activation values are related to the self-attention values in a transformer, as described in more detail by \citet{dai2021knowledge}.

Integrated gradients is a knowledge attribution method proposed by \citet{sundararajan2017axiomatic}, providing a method to evaluate the contribution of each neuron to knowledge predictions. The method involves gradually changing the value of a specific weight in the network while measuring the effect on the gradient of the model output calculated with respect to the weight of interest. If the given neuron is a knowledge neuron, its integration value will be low, given that the gradient value does not change upon perturbation. This method has proven effective for interpreting models in several application areas such as for diabetic retinopathy detection and machine translation \cite{sayres2019using}, \cite{sanyal2021discretized}.

Given that activation-based and integrated gradient-based interpretability methods are fairly new, they have not been applied to protein language models. However, a related class of interpretability methods is popular in the computational biology domain: attention weight analysis. For example, a group of researchers investigated attention in protein language models by iterating through the model’s layers, visualizing attention weights, and then analyzing the extent to which each one resembles a protein contact map \cite{vig2020bertology}. This method enables connections between certain model layers and biological concepts, but attention weight analysis does not yield importance values for all neurons in the attention head. Therefore, attention weight analysis is out of the scope of this project, though activation-based methods are motivated by the self-attention mechanism. In addition, interpreting a model by its activations or integrated gradients can offer benefits over attention weight analysis. For example, these methods capture information flow while attention weights do not \cite{qiang2022attcat}.

\subsection{Protein Language Models}

\begin{figure}[h]
\includegraphics[width=7cm]{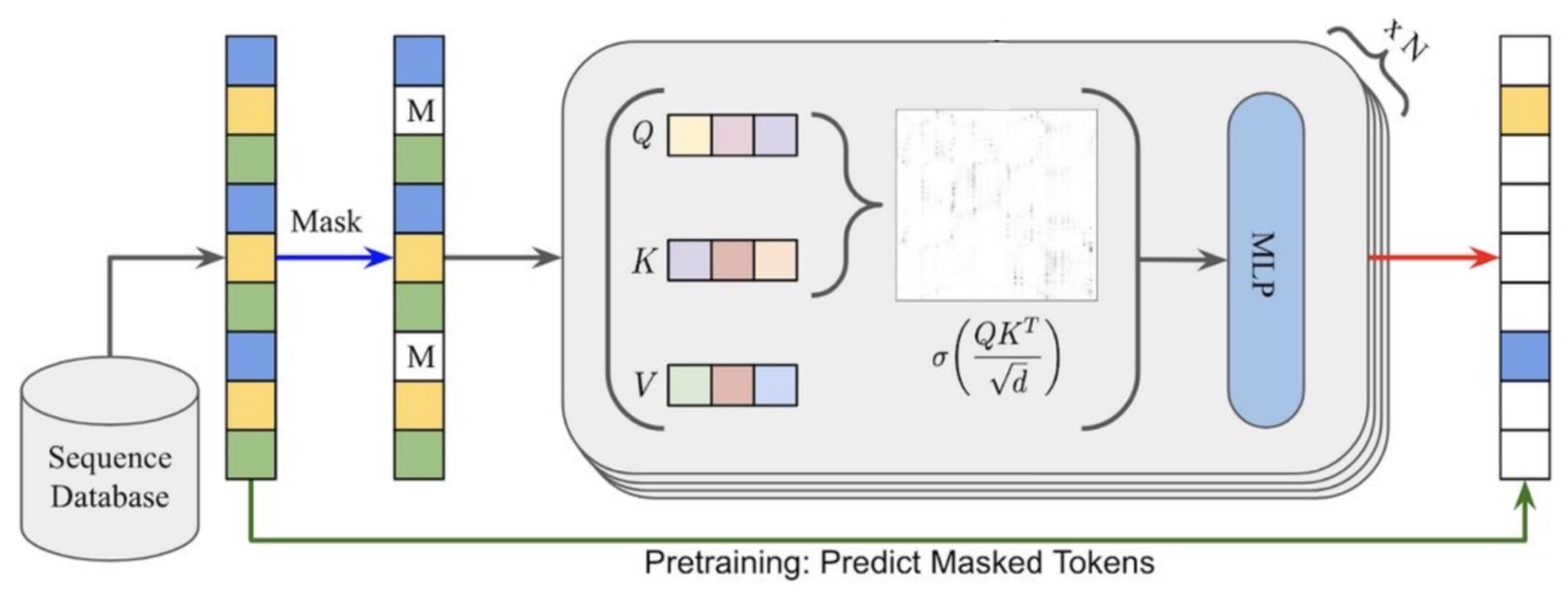}
\caption{ESM model architecture consists of several stacked attention heads, along with an NLP to predict the identity of masked amino acids.}
\label{esm}
\end{figure}

Protein language models are trained on large-scale protein sequence databases for the task of structure or function prediction \cite{hu2022protein}. They can be trained using a variety of deep learning architectures, such as LSTMs or transformers. The first examples of deep learning models trained on protein sequences became popular in 2018 when a convolutional neural network was trained on sequence features to predict structural properties \cite{jones2018high}. Over time, many modifications were made such as replacing the CNN with more sequence-suited architectures and adding multiple sequence alignment (MSA) information. In 2019, the first iteration of the ESM model was released, achieving state-of-the-art results \cite{rives2019biological}. As shown in Figure \ref{esm}, the ESM model is a transformer trained as a masked language model. The model consists of six self-attention heads, with several dropout layers in between. In our work, we use a base model of $8$ million parameters.

\section{Methods}

\subsection{Selected Task: Enzyme Classification}

As stated previously, here we focus on the interpretation of the ESM protein language model which is pre-trained on over $50,000$ protein sequences \cite{verkuil2022language}. We stack a linear classification head onto the encoder and fine-tune the full model for the task of enzyme sequence classification. From the Protein Data Bank (PDB), a set of $11,731$ enzyme sequences were downloaded, each with an enzyme class label \cite{PDB}. The six classes are oxidoreductases, hydrolases, transferases, lyases, isomerases, and ligases, and each enzyme class has roughly the same number of sequences in the dataset. This dataset was randomly split into $80\%$ train and $20\%$ test, and the training set was used to fine-tune the model. The test set was used for all evaluations.

We hypothesize that enzyme classification is a suitable task for knowledge neuron discovery for several reasons. First, enzymes often exhibit common motifs, or specific sub-sequences within the overall sequence, that are generally good indicators of being a certain enzyme and thus essential for accurate classification. Knowledge neurons could specialize in recognizing these motifs. Additionally, enzyme classification is a task that relies heavily on the evolutionary information learned by the pre-trained ESM encoder but does not rely on \textit{all} aspects of what ESM has been trained to learn. It is therefore plausible to hypothesize that a subset of the model can still perform well on this task.

\subsection{Experimental Set-Up}

We compare two potential methods to identify knowledge neurons in our trained enzyme classification language model, along with a baseline method. Similar to \citet{dai2021knowledge}, we hypothesize that important factual knowledge in pre-trained language models is stored in the query-key-value prediction feed-forward networks. Therefore, we focus on identifying knowledge neurons in the EsmSelfAttention modules contained in our model architecture.

Identifying knowledge neurons involves ablating or essentially "turning off" all other neurons to create a subset of the original model. We refer to this computed subset of neurons as a submodel. To ablate an input neuron of a linear layer, or effectively remove it from the model, all weights and biases associated with the neuron are set to zero.

We compare different knowledge neuron selection methods. First, our baseline method involves randomly selecting a subset of neurons, given a target percentage of parameters to preserve. We compare the effect of preserving an overall of $50\%$, $25\%$, $10\%$, and $1\%$ of parameters. Our next selection method is activation-based, in which we preserve neurons whose mean activation magnitudes on the test set are the same value or higher than a given percentile, again defined by a target percentage.

The final knowledge neuron selection method we assess is based on integrated gradients. From an intuitive perspective, as stated previously, the value of the integrated gradients measures the effect of perturbing a specific model weight. If the effect is significant, meaning that the output of the model highly depends on this weight's value, the weight is likely associated with a knowledge neuron. 

Here, we formally define our procedure. Given an input enzyme sequence $x$, the probability of the model predicting the correct enzyme class is given by the following equation:

\[P_x(\hat{w_i}^{(l)}) = p(y^* | x, w_i^{(l)} = \hat{w_i}^{(l)} )\]

In this equation, $w_i^{(l)}$ is the $i$th neuron in the $l$th attention head and $y^*$ denotes the correct answer. $\hat{w_i^{l}}$ is the value that the corresponding weight has been set to after training. To calculate the integrated gradient, we follow this equation as given by the Captum library \cite{Captum}.

\[IG(w_i^{(l)}) = \bar{w_i}^{(l)} (\int_{\alpha=0}^{1} \frac{\partial P_x(\alpha \bar{w_i}^{(l)})}{\partial w_i^{(l)}} \, d\alpha) \]

The partial derivative calculates the gradient of the model output with respect to the neuron weight of interest. Since directly calculating the continuous integral is intractable, in the Captum library, the integral is calculated as a Reimann sum.

For each neuron weight, integrated gradients were calculated with respect to every data point in the test set's corresponding output. We take the average over the test set, and average over the weights, to arrive at an importance value for each neuron. This process was highly memory-intensive, so it was conducted once on a cloud GPU cluster, and then serialized for further analysis.

\subsection{Evaluation}

To evaluate the efficacy of each method in their ability to identify knowledge neurons, we construct submodels comprised of the identified knowledge neurons and measure the submodels' accuracy on the test set. This analysis is conducted at various submodel sizes, as a percentage of the original model size. Specifically, we construct submodels containing $50\%$, $25\%$, and $10\%$, and $1\%$ of self-attention head neurons. Additionally, we visualize the knowledge neurons with respect to the overall model architecture. 

\section{Results}

\subsection{Activations and Gradients Identify Knowledge Neurons}

Using the random selection method, activation-based selection method, and integration gradient-based selection method, $4$ different sizes of submodels were created, ranging from preserving $50\%$ to $1\%$ of the original model's self-attention module parameters. Ideally, these submodels contain all predicted knowledge neurons. Figure \ref{fig:1} displays the accuracy of these submodels on the test set of enzyme sequences, and Table \ref{tab:1} shows the exact test accuracy values in each setting.

\begin{table*}[h]
\centering
\begin{tabular}{llllll}
\hline
\textbf{Methodology} & \textbf{100\%}& \textbf{50\%} & \textbf{25\%} & \textbf{10\%} & \textbf{1\%}\\
\hline
\verb|Random Selection| & 0.9131 & 0.1798 & 0.0878 & 0.1866 & 0.1866 \\
\verb|Activation-Based Selection| & 0.9131 & 0.2850 & 0.2842 & 0.2889 & 0.2965 \\
\verb|Integrated Gradients-Based Selection| & 0.9131 & 0.3550 & 0.3122 & 0.2890 & 0.2965 \\ \hline
\end{tabular}
\caption{We compare the efficacy of activation-based and integrated gradients-based selection of knowledge neurons and compare to random selection. The columns denote the size of the submodel, comprised of predicted knowledge neurons, as a percentage of the original model's size.}
\label{tab:1}
\end{table*}

\begin{figure}[h]
\includegraphics[width=7cm]{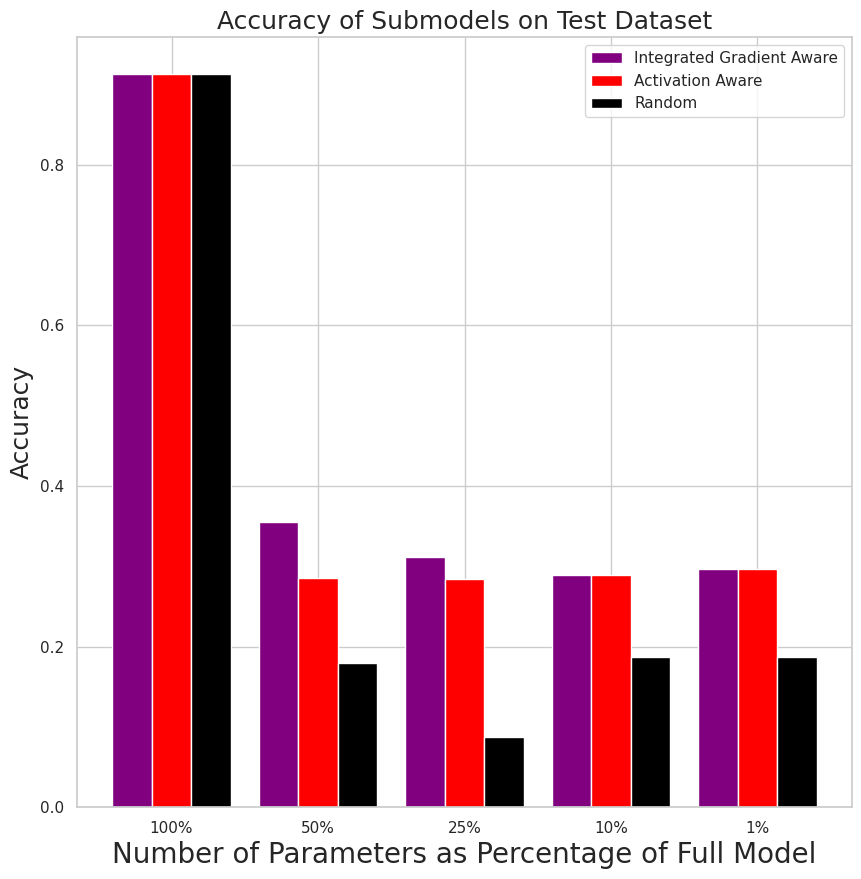}
\caption{Bar graph displaying the test accuracy of various submodels on the task of enzyme sequence classification. The black bars denote performance of submodels created by random selection. The red bars correspond to activation aware selection and the purple bars correspond to integrated gradient aware selection.}
\label{fig:1}
\end{figure}

\begin{figure*}[ht]
    \centering
    \includegraphics[width=2.25\columnwidth]{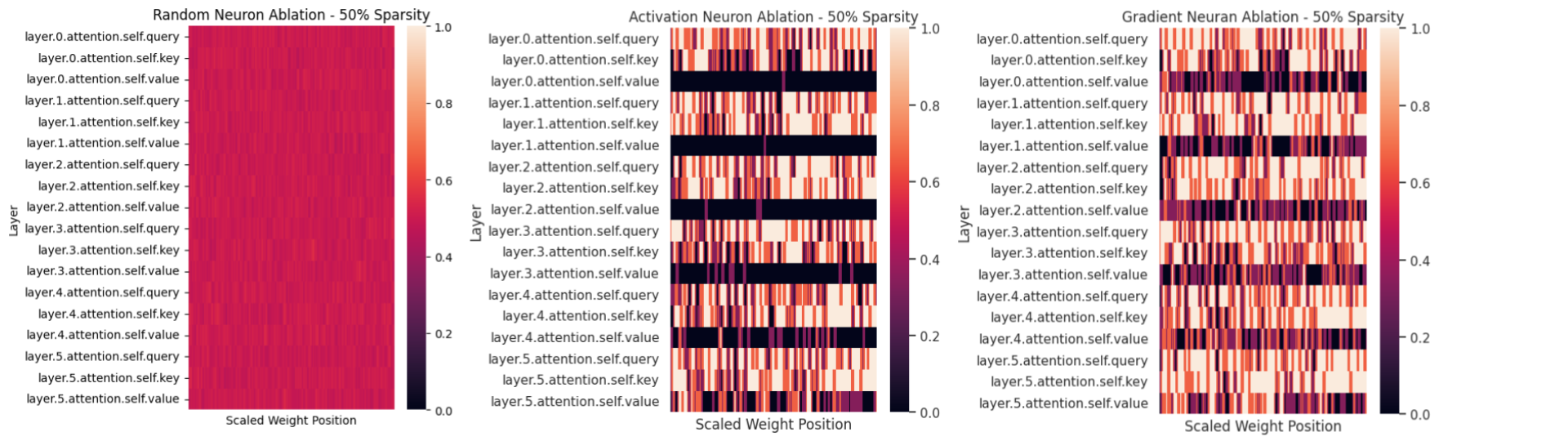}
    \caption{From these heatmaps, we can see that when using 50\% sparsity, neurons related to the value are the least important for knowledge expression. When using random neuron ablation, as demonstrated in the first heatmap, there is no discernable pattern in the neurons that are dropped. However, in both activation neuron ablation (second heatmap) and gradient neuron ablation (third heatmap), we can see that there are darker bars across the layers labeled as values. This means that these layers are dropped using the respective algorithm, indicating that these neurons are less important in knowledge expression. Therefore, from using 50\% sparsity, we can conclude that the neurons related to value are not as important for knowledge expression.}
    \label{fig:1}
\end{figure*}

\begin{figure*}[ht]
    \centering
    \includegraphics[width=2.25\columnwidth]{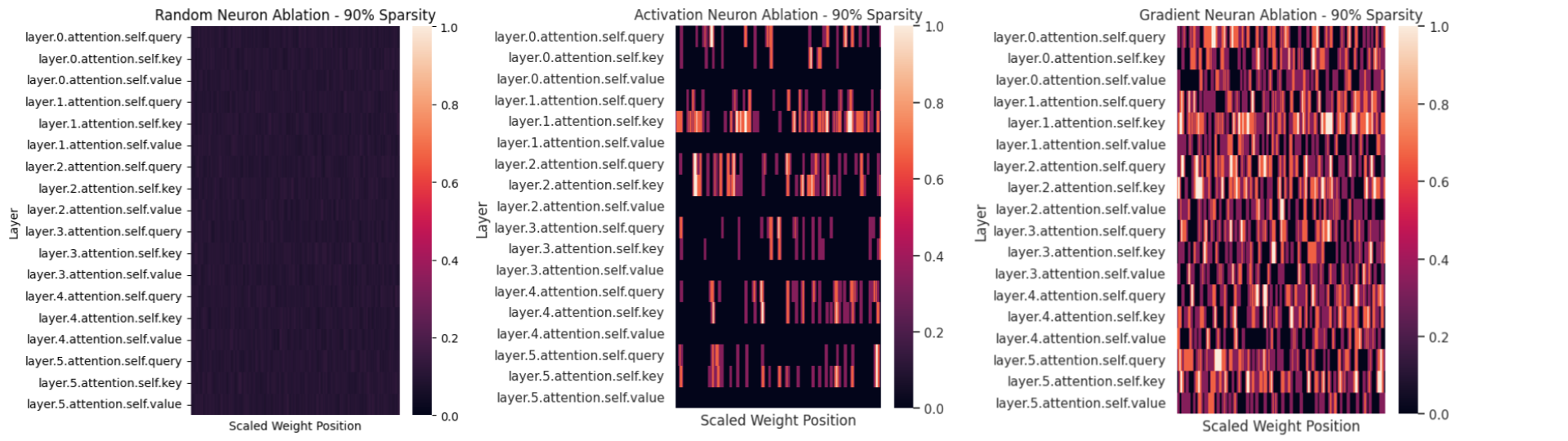}
    \caption{As sparsity increases to 90\%, the activation neuron ablation starts identifying knowledge expression in all layers while the gradient still primarily sticks to value neurons. This indicates that the gradient neuron ablation might be the best approach for determining which neurons to drop.}
    \label{fig:2}
\end{figure*}

\begin{figure*}[ht]
    \centering
    \includegraphics[width=2.25\columnwidth]{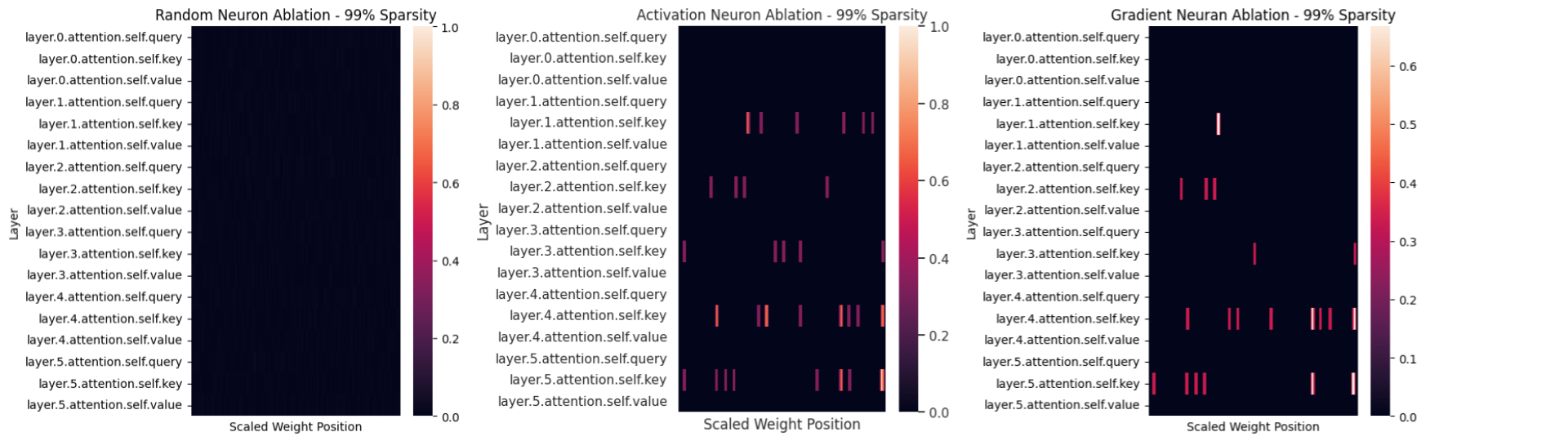}
    \caption{We can see that the key neurons are the most important for knowledge expression. In the random neuron ablation (first heatmap) there doesn't seem to be an apparent pattern. However, with both activation neuron ablation and gradient neuron ablation, we can see that the layers with key neurons seem to be the only ones that are not completely purple and have other colors in them. This suggests that the only type of neurons that don't seem to be dropped are the ones related to keys, indicating that the key neurons are the most important for knowledge expression.}
    \label{fig:3}
\end{figure*}

The test accuracy of the original model is $0.9131$. Given the drastic drop in accuracy between the original model and even the largest submodel, it is clear that many of the query, key, and value neurons in the pre-trained ESM model are very important. However, given the notable difference we observe between the random selection and activation-based and gradient-based selection methods, some neurons are evidently more important than others. These neurons can be considered knowledge neurons, though they are not responsible for learning \textit{all} relevant knowledge. 

We observe that the integrated gradient aware selection method performs the best for the $50\%$ and $25\%$ submodels. The activation and gradient methods perform almost the same for the smaller submodels of sizes $10\%$ and $1\%$. From this result, we hypothesize that when asked to select fewer knowledge neurons, the activation and integrated gradients method may reach a consensus. When asked to select a larger number of knowledge neurons, the integrated gradients method selects a subset of neurons with higher knowledge expression.

Overall, given that submodels created by activation and integrated gradient-based selection consistently outperform the random baseline, we can conclude that these methods successfully identify knowledge neurons.

\subsection{Analysis of Identified Knowledge Neurons}

We then perform further analysis on submodels of $50\%$, $90\%$, and $99\%$ sparsity, equivalent to preserving  $50\%$, $10\%$, and $1\%$ of neurons. In Figure \ref{fig:1}, we show analysis at $50\%$ sparsity. To create these heatmaps, each neuron in the original model was labeled with a $0$ or $1$, corresponding to an ablated neuron or knowledge neuron respectively. Each cell in the heatmap corresponds to a group of $200$ adjacent neurons, and the displayed value is the average label of neurons in the group. Therefore, darker values indicate low density of larger neurons. Additionally, rows in the heatmap correspond to different layers, as labeled.

In Figure \ref{fig:1}, we observe that with activation-aware ablation and gradient-aware ablation, neurons in the value vector prediction feed-forward networks are the least important. When using random neuron ablation, as demonstrated in the first heatmap, neurons are uniformly dropped across all layers. Therefore, there are no discernible patterns, and all heatmap values are approximately $0.5$. However, in both activation neuron ablation (center heatmap) and gradient neuron ablation (right-most heatmap), we can see that there are darker bars across the layers labeled as values. This means that many neurons in these layers are dropped, indicating that the neurons in these layers are less important for knowledge expression. Therefore, at $50\%$ sparsity, we can conclude that neurons in key and query prediction networks are more important.

Figure \ref{fig:2} shows the same analysis for $90\%$ sparsity. As sparsity increases to $90\%$, the integrated gradients method starts identifying knowledge expression in all layers while the activation-aware method still primarily sticks to key and query neurons. Finally, in Figure \ref{fig:3}, we see that both the activation and integrated gradients methods select neurons in the key vector prediction networks.

Here, we hypothesize why the key-related neurons are predicted to be most important for knowledge expression. Placing important neurons in the key prediction FFNs could allow for a more selective attention mechanism, meaning that the model can focus on specific aspects of the input sequence by assigning higher importance to certain key vectors. Given that key vectors represent different features of the input sequence, it is likely that these knowledge neurons specialize in capturing knowledge of different enzyme sequence motifs. Additionally, key vectors often serve as a compressed representation of the input sequence, and performing this dimensionality reduction step in a way that aids the task is important.

\section{Conclusions}

In this research, we addressed the challenge of interpretability in protein language models, focusing on the Evolutionary Scale Model (ESM-2). Our goal was to identify knowledge neurons within the model, shedding light on the components that contribute most significantly to its predictions in the context of enzyme sequence classification. We employed activation-based and integrated gradients-based methods to identify knowledge neurons, comparing their performance against a random selection baseline. Our results demonstrate that both activation and integrated gradients methods consistently outperform random selection, indicating their effectiveness in pinpointing neurons crucial for knowledge expression. Specifically, the integrated gradients method performed the best, with activation-aware selection following closely after. Moreover, our analysis revealed intriguing patterns in the importance of different types of neurons. Specifically, neurons related to key vectors in the self-attention mechanism were consistently identified as crucial for knowledge expression. This finding suggests that these knowledge neurons specialize in understanding different parts of the input enzyme sequence.

The knowledge gained from this research can guide further improvements in model architectures and training strategies, potentially leading to more accurate and reliable protein language models. For example, for the task of enzyme sequence classification, increasing the dimensionality of key vectors in the ESM encoder may improve performance, given that the key neurons are learning crucial information.

While our current research provides valuable insights, there are avenues for future exploration. One promising direction is to delve deeper into knowledge neuron analysis by moving beyond binary classification (knowledge neuron vs ablated neuron) and instead classifying the types of knowledge encoded by specific neurons. This could involve developing methods to categorize the information captured by neurons into distinct biological or structural features. Understanding the nuanced nature of knowledge representation in protein language models could provide more fine-grained insights into the model's comprehension and decision-making processes. Additionally, extending the analysis to different tasks within computational biology and applying these interpretability methods to other state-of-the-art protein language models would further validate and generalize our findings.

\section{Impact Statement}

Here, we highlight the impact of our work on the existing state of machine learning applications and society at-large. Protein language models are increasingly being used for real-world drug discovery applications. For example, ESM protein embeddings are frequently used to represent protein binding targets for downstream molecular generation tasks. Generative models use the ESM embedding to design small molecule drugs that bind to the given protein. 

On one hand, further work on protein language model interpretability could enhance the quality of such embeddings. For example, as we did in this work, understanding the most important components of the pre-trained ESM model for enzyme sequence classification could lead to architecture optimizations for this task. This would then lead to higher quality enzyme vector representations, allowing a downstream generative model to design enzyme inhibitor molecules for therapeutic applications with higher success rate.

On the other hand, it is important to note that the ESM model, and related biological language models, should not be applied blindly. Though our work aims to shed light on the model's inner workings, it is still difficult to interpret the specific knowledge each neuron learns. Therefore, human experts are required to verify all predictions.

It is also important to acknowledge and address possible biases in our study. We acknowledge that our study is heavily based on a pre-trained language model, ESM, which may encode biases from its training data and training procedure. Particularly, the ESM model is trained and fine-tuned on a subset of the Protein Data Bank, a dataset where proteins with healthy function are most represented. Therefore, applying the findings of this study to inform tasks dealing with disordered proteins should be done very carefully. In our study, we also ensured we worked with only publicly available, non-personally identifying data that does not put any individual at privacy risk. 

Based on this analysis, we recommend that biologists and clinical experts work together with machine learning researchers to pave a path towards useful interpretation of biological language models. We hope this work sheds light on the importance of interpretability research, motivating the field to collectively take steps towards understanding the "black box" of LMs. 

\bibliographystyle{acl_natbib}
\bibliography{custom}

\end{document}